\documentclass{article}
\usepackage{spconf,amsmath,graphicx}

\usepackage{enumitem}
\setlist{nosep, leftmargin=14pt}

\usepackage{subfig}
\usepackage{tabulary}
\usepackage{multirow}
\usepackage{booktabs}
\usepackage{colortbl}
\usepackage{tabularx}
\usepackage{multirow}
\usepackage{threeparttable}
\usepackage{colortbl}
\usepackage{amssymb}
\usepackage{latexsym}

\usepackage{hyperref}
\usepackage{url}
\usepackage{titlecaps}

\usepackage{multicol}
\usepackage{graphicx}
\usepackage{xcolor,cancel,mathrsfs,amscd}
\usepackage{subfig}
\usepackage{amsmath}
\usepackage{mfirstuc}
\title{How precise are performance estimates for typical medical image segmentation tasks?}

\name{Rosana El Jurdi, Olivier Colliot}
\address{Sorbonne Université, Institut du Cerveau - Paris Brain Institute - ICM, CNRS, Inria, Inserm, AP-HP,\\ Hôpital de la Pitié Salpêtrière, F-75013, Paris, France}

\begin{document}
%
\maketitle
\begin{abstract}
An important issue in medical image processing is to be able to estimate not only the performances of algorithms but also the precision of the estimation of these performances. Reporting precision typically amounts to reporting standard-error of the mean (SEM) or equivalently confidence intervals. However, this is rarely done in medical image segmentation studies. In this paper, we aim to estimate what is the typical confidence that can be expected in such studies. To that end, we first perform experiments for Dice metric estimation using a standard deep learning model (U-net) and a classical task from the Medical Segmentation Decathlon. We extensively study precision estimation using both Gaussian assumption and bootstrapping (which does not require any assumption on the distribution). We then perform simulations for other test set sizes and performance spreads. Overall, our work shows that small test sets lead to wide confidence intervals (e.g. $\sim$8 points of Dice for 20 samples with $\sigma \simeq 10$). 
\end{abstract}

\begin{keywords}
Segmentation, Performance, Validation, Statistical analysis, Confidence interval, Standard error.
\end{keywords}
\section{Introduction}
\label{sec:intro}

In medical imaging, it is not uncommon that sample sizes are in the order of dozens of subjects, at best hundreds or thousands. In 3D medical image segmentation, the size of the set used to evaluate the performance may be even smaller than for other medical imaging tasks as obtaining the ground truth requires voxel-wise annotation by trained raters.

Intuitively, the precision of the estimation of the performance depends on two factors: the variability of the performance among the test set (the more variable, the less precise) and the size of the test set (smaller sets will lead to lower precision and therefore larger confidence intervals). However, papers usually report the average performance for different metrics (e.g. average Dice) but not the precision~\footnote{Throughout the paper, precision means how precise are the estimates of the performance. It has nothing to do with the performance metric Precision also known as Positive Predictive Value.} with which this average performance is estimated. Such precision can be provided in the form of confidence intervals or equivalently standard error of the mean (SEM) which are not often reported. What is more often reported is the empirical standard deviation over different folds of a cross-validation. While this may qualitatively characterize the variability of the learning procedure when the training and testing set change, it should never be used to compute the SEM, since here $n$ would be the number of folds or splits, which is arbitrary and can be made as large as one wants, thereby making the confidence interval arbitrarily narrow. It is not even an unbiased estimate of the standard deviation of the performance metric~\cite{bengio2004no}.

Quantifying the precision of the estimation of the performances thus requires an independent test set, on which confidence intervals or SEM are reported. Since this is not typically done in medical image segmentation papers, one may ask the following question. What precision can be expected for a typical sample size? How trustworthy are the average performance estimates (for instance Dice coefficients) reported in medical image segmentation papers? 

Surprisingly, this question has  been little studied in medical imaging. In the case of a different task, namely image classification, it is necessary to have large sample sizes for a precise estimation of the accuracy (typically 10,000 samples to achieve a $1\%$-wide confidence interval given an accuracy of about $90\%-95\%$)~\cite{VAROQUAUX201868,varoquaux2022}. 
However, to the best of our knowledge, this is not widely known in the case of segmentation. We hypothesize that the test size needed to achieve a given precision is lower than for classification due to the continuous nature of performance measures~\cite{Altman2006}. 

Our objective is to study the precision that can be expected in 3D medical image segmentation for typical test set sizes. We first conduct experiments using a standard deep learning network applied  to a classical segmentation task from the Segmentation Decathlon Challenge~\cite{decathlon_short} in order to estimate confidence intervals which are obtained for variable test set sizes. We then perform simulations for other sizes and spreads. We insist that the aim of the present paper is not to propose a new segmentation methodology. Instead, the main aims are to provide information regarding the confidence intervals that can typically be expected in medical image segmentation and to raise awareness of the community on this important issue.

\section{Overview}
\label{sec:overview}

Our aim is to provide confidence intervals (or standard-errors) for the mean of a given performance metric for different test set sizes and different spreads (standard deviation) of the performance. If the performance metric follows a Gaussian distribution, one can obtain those values as follows:
\begin{align}
\begin{split}
 &    \text{SEM} = \frac{\sigma}{\sqrt{n}} \\ 
 &  \text{CI} = [\mu-1.96\times \text{SEM},\mu+1.96 \times \text{SEM}]
    \label{eqn:gaussian_assumption}
\end{split}
\end{align}

where $\mu$ denotes the mean, $\sigma$ the standard-deviation, $n$ the test set size, $\text{SEM}$ the standard error of the mean and $\text{CI}$ the 95\% confidence interval. In the following, for conciseness, we will also denote the width of $\text{CI}=[a,b]$ as $w=b-a$.

One does not know a priori if using Equations~\ref{eqn:gaussian_assumption} is valid for a given performance metric. On the other hand, in the absence of assumption on the distribution of the metric, one can use bootstrapping to estimate $\text{SEM}$ and $\text{CI}$~\cite{efron1994introduction}. In Section~\ref{sec:Experiments-hippo}, we will  perform experiments in order: i) to verify the validity of using Equations~\ref{eqn:gaussian_assumption} by comparing the estimates obtained using these equations to those using bootstrapping; ii) obtain empirical estimates of $\mu$ and $\sigma$; iii) study the effect of using subsamples of reduced size.

In Section~\ref{sec:simulation}, we will perform simulations using Equations~\ref{eqn:gaussian_assumption} to study how the precision varies when varying the test set size and performance variability (as measured by $\sigma$).

\begin{table*}[t]
 \centering
    \begin{tabular}{|l|cccc|ccc|}
    \hline
n & $\mu$ & $\sigma$ & $SEM$ & $w$ & $\mu^*$ & $SEM^*$ & $w^*$ \\ \hline
n = 110 & 80.70&10.75&1.02&4.02&80.70&1.02&3.99\\\hline

    \end{tabular}
    \vspace{0.2cm}
    \caption{\textbf{Results on the full test set ($ n = 110$)}. $\mu$ and   $\sigma$  are the empirical mean and standard deviation of the Dice coefficient across all patients in the test set. $SEM$ is the standard error of the mean  and $w$ is the width of the $95\%$ confidence interval calculated via Equations~\ref{eqn:gaussian_assumption}. $SEM^*$, $\mu^*$ and $w^*$ are the values obtained via Bootstrapping.}
    \label{tab:Stats_Hippo_110}
\end{table*}

\section{Experiments}
\label{sec:Experiments-hippo}

\subsection{Dataset and segmentation method}

We used the Hippocampus dataset from the Medical Decathlon challenge~\cite{decathlon_short}, composed of 260 3D  MR images.
The task is to segment the anterior and posterior parts of the hippocampus. For evaluation, we merged the two regions (in prediction and ground-truth respectively) and considered the hippocampus as a whole.
From the 260 samples, 100 patients were randomly selected for training, 50 for validation and the remaining 110 samples constituted the test set. 

For the segmentation, we used a U-net type network~\cite{ronneberger2015u}. Note that our aim is not to achieve the highest possible segmentation scores  or to propose a novelty in the segmentation method but rather to obtain {\sl typical} performances. We thus relied on a standard approach. We treat a 3D MRI as a sequence of 2D images and used a 2D architecture. At inference, we predict for each slice independently then stack the slices belonging to the same patient together to form a 3D-volume prediction. The architecture is a 3-stage structure composed of convolutional and de-convolutional blocks, bottleneck and skip connections. An ensemble of convolutional and batch normalization layers constitutes the encoder part. Each stage within the decoder path is composed of 2 consecutive convolutional blocks followed by an upsampling layer. The bottleneck is composed of 2 convolutional blocks separated by a residual block~\cite{Zhang2018a}. The architecture has been used in previous publications~\cite{Kervadec2019a, jurdi2021a}. The optimizer was Adam, the learning rate was $0.001$ 
and the batch size was 8. The learning rate was halved if the validation performances
did not improve over 20 epochs as proposed by ~\cite{Kervadec2019a}.
Note that we used the standard generalized Dice loss \cite{Sudre2017}. The model was trained over 500 epochs. We performed a three-fold cross-validation with the 150 patients of the training/validation sets and selected the best model across these three folds. Note that the test set was left untouched and was never used at any stage for training, model selection or architecture/parameter optimization.  To evaluate the performance, we computed the Dice coefficient in $\% $~\footnote{Performance metric was computed using this code: \url{https://github.com/deepmind/surface-distance}} for the whole hippocampus (thus merging anterior and posterior parts before computing the metric). 

The code used to generate the results is available online 
\footnote{\footnotesize
 \url{ https://github.com/rosanajurdi/SegVal/tree/ISBI2023}
}

\begin{figure}[h!]
\vspace{-0.2cm}
	   \includegraphics[width=9cm,height=7cm]{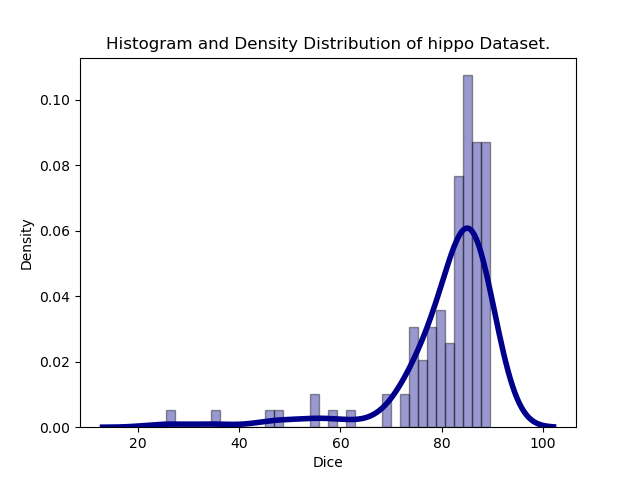}   
	   \label{Dist_Hippo_total}
\vspace{-0.7cm}
\caption{
Histogram of Dice accuracy over the entire test set. It is shown together with a kernel density estimation (KDE) which smoothes the observations with a Gaussian kernel.}
\label{fig:Hippo_110_Distribution}
\end{figure}

\begin{table*}[t!]
 \centering
    \begin{tabular}{|l|cccc|ccc| }
\hline    
Subsample size $k$ & $\mu_k$  & $\sigma_k$& $\text{SEM}_k$ & $w_k$ & $\mu^*_k$  &  $\text{SEM}^*_k$ & $w^*_k$ \\ \hline
$k$ = 10&81.01 $\pm$ 3.04 &8.17 $\pm$ 4.75&2.58 $\pm$ 1.5&10.13 $\pm$ 5.88 & 81.01 $\pm$ 3.04 &2.59 $\pm$ 1.51&9.86 $\pm$ 5.62\\
 $k$= 20&80.61 $\pm$ 2.16 &9.96 $\pm$ 3.74&2.23 $\pm$ 0.84&8.73 $\pm$ 3.28 &80.61 $\pm$ 2.16 &2.23 $\pm$ 0.84&8.61 $\pm$ 3.21\\
$k$= 30&80.63 $\pm$ 1.6 &10.36 $\pm$ 2.94&1.89 $\pm$ 0.54&7.41 $\pm$ 2.1 &80.64 $\pm$ 1.6 &1.89 $\pm$ 0.54&7.34 $\pm$ 2.08\\
 $k$= 50&80.95 $\pm$ 1.14 &9.99 $\pm$ 2.1&1.41 $\pm$ 0.3&5.54 $\pm$ 1.16 & 80.95 $\pm$ 1.14 &1.41 $\pm$ 0.3&5.51 $\pm$ 1.15 \\
$k$= 100&80.64 $\pm$ 0.32 &10.78 $\pm$ 0.53&1.08 $\pm$ 0.05&4.22 $\pm$ 0.21 & 80.64 $\pm$ 0.31 &1.08 $\pm$ 0.05&4.21 $\pm$ 0.21\\
$k$= 110&80.70 $\pm$ 0.0 &10.75 $\pm$ 0.0&1.02 $\pm$ 0.0&4.02 $\pm$ 0.0 & 80.71 $\pm$ 0.01 &1.02 $\pm$ 0.01&3.99 $\pm$ 0.03 \\
\hline
    \end{tabular}
        \vspace{0.2cm}
 \caption{\textbf{Results on subsamples of size $k \leq 110 $.} Results are shown as $mean \pm std$ where $mean$ and $std$ are the mean and standard-deviation over all the subsamples $S_{k,j}$ of a given size  $k$ ($k$ is fixed and  $ j \in \{1, \ldots, 100\}$).}
    \label{tab:hippo_subsamples_k}
\end{table*}

\subsection{Precision on the whole test set}

We first studied the precision of performance estimates using the maximum test set size.  The distribution of Dice values over the test set is shown in \figureautorefname~\ref{fig:Hippo_110_Distribution}. One can observe that the Gaussian assumption is not unreasonable despite underlying outliers and skewness. More importantly, we will now compare estimates based on this assumptions to corresponding non-parametric bootstrap estimates.

We first compute $\mu$, $\sigma$, $\text{SEM}$ and $w$ using Equations~\ref{eqn:gaussian_assumption}. We then compute their bootstrap counterparts $\mu^*$, $\text{SEM}^*$, $w^*$ as follows~\footnote{\footnotesize Throughout the paper, the bootstrap estimate of a given $x$ is always denoted as $x^*$}. 
Given a test set of size $n$, $ M=15000$ bootstrap samples of size $n$ are drawn  with replacement. We denote  a given bootstrap sample as $S^*_m$ and its mean as $\mu_m^*$ where $m \in \{ 1, \ldots, M \}$. The bootstrap mean $\mu^*$ is the mean of the bootstrap sample means $\mu_m^*$. The standard error of the mean $\mu^*$ obtained via bootstrapping ($\text{SEM}^*$) is the standard deviation of the means of all bootstrap samples:
$
\text{SEM}^* = \sqrt{\frac{1}{M} \sum_{m=1}^{M}{\left(\mu_m^*-\mu^*\right)^2}}
$. The $95\% $ confidence interval $\text{CI}^*=[a^*, b^*]$ is the set of values between the $2.5 \% $ and $97.5\%$ percentiles of the sorted bootstrap means $ \{\mu_1^*,\mu_2^*, \ldots, \mu_m^*,\ldots \mu_M^*\}$. We finally define the width of the confidence interval as $w^* = b^* - a^*$. The estimates using the Gaussian assumption and the bootstrap are very close as can be seen on \tableautorefname ~\ref{tab:Stats_Hippo_110}.

\subsection{Precision on subsamples of size $k \leq n= 110$}

We now study experimentally the relationship existing between the test set size and the precision of the estimation of the segmentation performance.
To that end, we draw subsamples of variable size $ k \in K=\{10,20,30,50,100,110\}$ from the whole test set of size $n$. In order not to depend on a particular drawing (which may be lucky or unlucky), we repeat the procedure 100 times for each $k$. We denote the subsamples as $(S_{k,j})$ where $k$ is the subsample size and $ j \in \{1, \ldots , 100\}$ is the index of a particular drawing. 

We then proceed with the computations of the different estimates based either on the Gaussian assumption or on the bootstrap.

For the Gaussian assumption, we denote as  $\mu_{k,j}=\frac{1}{k} \sum_{l=1}^{k}D_{j,k,l}$ and  $\sigma_{k,j}=\sqrt{\frac{1}{k} \sum_{l=1}^{k}{\left(D_{j,k,l}-\mu_{k,j}\right)^2}}$ the empirical mean and standard deviation for the subsample $S_{k,j}$ where $D_{j,k,l}$ is the Dice coefficient of a given subject in the subsample $S_{k,j}$. Similarly, we use the notations $\text{SEM}_{k,j}=\frac{\sigma_{k,j}}{\sqrt{k}}$ and  $w_{k,j}=2*1.96*\text{SEM}_{k,j}$.
We can then study how these values vary across the 100 subsamples of a given size $k$. To that end, we compute the average and standard-deviation of $\mu_{k,j}$, $\sigma_{k,j}$, $\text{SEM}_{k,j}$ and $w_{k,j}$ across the different subsamples $S_{k,j}$ for $k$ fixed and  $ j \in \{1,\ldots, 100\}$. This provides the following  estimates $\mu_{k}\pm\sigma_{\mu_{k}}$, $\sigma_{k}\pm\sigma_{\sigma_{k}}$, $\text{SEM}_{k}\pm\sigma_{\text{SEM}_{k}}$ and $w_{k}\pm\sigma_{w_{k}}$. The values are displayed in Table~\ref{tab:hippo_subsamples_k}. 
One can gather that, as the sample size increases, the standard deviation and the standard error decrease. 

\begin{table*}[htb]
 \centering
\begin{tabular}{|l|cc|cc|cc|cc|cc|cc|cc|}
\hline

$\sigma$  &   \multicolumn{2}{c}{2} \vline &   \multicolumn{2}{c}{5} \vline &   \multicolumn{2}{c}{8} \vline & \multicolumn{2}{c}{\cellcolor{lightgray}10.75} \vline &   \multicolumn{2}{c}{12} \vline &  \multicolumn{2}{c}{15} \vline & \multicolumn{2}{c}{18}\vline \\\hline
  & $\text{SEM}$ & $w_k$  & $\text{SEM}$ & $w_k$ & $\text{SEM}$ & $w_k$& $\text{SEM}$ & $w_k$& $\text{SEM}$ & $w_k$& $\text{SEM}$ & $w_k$ &$\text{SEM}$ & $w_k$\\
 \hline
$k = 10$&0.63&2.48&1.58&6.2&2.53&9.92&\cellcolor{lightgray}3.4&\cellcolor{lightgray}13.33&3.79&14.88&4.74&18.59&5.69&22.31\\ 
$k = 20$&0.45&1.75&1.12&4.38&1.79&7.01&\cellcolor{lightgray}2.4&\cellcolor{lightgray}9.43&2.68&10.52&3.35&13.15&4.02&15.78\\ 
$k = 30$&0.37&1.43&0.91&3.58&1.46&5.73&\cellcolor{lightgray}1.96&\cellcolor{lightgray}7.7&2.19&8.59&2.74&10.74&3.29&12.88\\ 
$k = 50$&0.28&1.11&0.71&2.77&1.13&4.43&\cellcolor{lightgray}1.52&\cellcolor{lightgray}5.96&1.7&6.65&2.12&8.32&2.55&9.98\\ 
$k = 100$&0.2&0.78&0.5&1.96&0.8&3.14&\cellcolor{lightgray}1.08&\cellcolor{lightgray}4.22&1.2&4.7&1.5&5.88&1.8&7.06\\ 
$k = 200$&0.14&0.55&0.35&1.39&0.57&2.22&\cellcolor{lightgray}0.76&\cellcolor{lightgray}2.98&0.85&3.33&1.06&4.16&1.27&4.99\\ 
$k = 300$&0.12&0.45&0.29&1.13&0.46&1.81&\cellcolor{lightgray}0.62&\cellcolor{lightgray}2.43&0.69&2.72&0.87&3.39&1.04&4.07\\
$k = 500$&0.09&0.35&0.22&0.88&0.36&1.4&\cellcolor{lightgray}0.48&\cellcolor{lightgray}1.89&0.54&2.1&0.67&2.63&0.8&3.16\\ 
$k = 1000$&0.06&0.25&0.16&0.62&0.25&0.99&\cellcolor{lightgray}0.34&\cellcolor{lightgray}1.33&0.38&1.49&0.47&1.86&0.57&2.23\\ 

\hline
  \end{tabular}
  \caption{Simulation of $\text{SEM}_k$ and the width $w_k$  for different sizes $k$ of the test set and for different values of $\sigma$. The gray column with $\sigma = 10.75$ corresponds to the  standard deviation found in the experimental section. }
  \label{gaussian}
\end{table*}

Bootstrap estimations are performed as follows.
 For a given subsample $S_{k,j}$ of size $k$ and index $j$, $M=15000$ bootstrap samples of size $k$ are drawn with replacement. We denote a given bootstrap sample as $S^*_{k,j,m}$ and its mean as $\mu^*_{k,j,m}$ where $m \in \{ 1, \ldots, M\}$ is the index of the $m^{th}$ bootstrap sample of subsample $S_{k,j}$. The bootstrap mean $\mu^*_{k,j}$ of $S_{k,j}$ is the mean of the bootstrap sample means   $\mu^*_{k,j,m}$: $ \mu^*_{k,j} = \frac{1}{M} \sum_{m=1}\mu^*_{k,j,m} $. The standard error of the mean $\mu^*_{k,j}$ (denoted as $\text{SEM}^*_{k,j}$) obtained via bootstrapping  is the standard deviation of the means of all bootstrap samples of subsample $S_{k,j}$: $
\text{SEM}^*_{k,j} = \sqrt{\frac{1}{M} \sum_{m=1}^{M}{\left(\mu^*_{k,j,m}-\mu^*_{k,j}\right)^2}}
$.  The $95\%$ confidence interval 
of the sample $S_{k,j}$ is denoted as $[a^*_{k,j}, b^*_{k,j}]$ and is the set of values between the $2.5 \% $ and $97.5\%$ percentile of the sorted bootstrap means of subsample $S_{k,j}$. We finally define the width of the confidence interval via bootstrapping as $w^*_{k,j}=b^*_{k,j}-a^*_{k,j}$. 
We study how these values vary across the 100 samples of a given size $k$. To that end, we compute the averages and the standard deviations of $\mu^*_{k,j}$,  $\text{SEM}^*_{k,j}$ and $w^*_{k,j}$ across the different subsamples $S_{k,j}$ for $k$ fixed and  $ j \in \{1, ..., 100\}$. This provides the following  estimates $\mu^*_{k}\pm\sigma_{\mu^*_{k}}$, $\text{SEM}^*_{k}\pm\sigma_{\text{SEM}^*_{k}}$ and $w^*_{k}\pm\sigma_{w^*_{k}}$. Results are shown in Table~\ref{tab:hippo_subsamples_k}. 

As for the whole test set, estimates using Equation~\ref{eqn:gaussian_assumption} and bootstrapping are very close across different subsample sizes. As expected, precision decreases with the sample size.

\section{Simulations with a Gaussian Distribution}
\label{sec:simulation}

In the previous section, we showed that estimates of $\text{SEM}$ and confidence intervals computed using either Equations~\ref{eqn:gaussian_assumption} or the bootstrap are in accordance. Given this, we now perform simulations using Equations~\ref{eqn:gaussian_assumption} for more values of $k$ (including in particular larger test sets) and for different values of  $\sigma$. 
Note that this is independent of $\mu$ which, in itself, has no impact  on the $\text{SEM}$ nor on the width of the CI (even though it is usually observed that lower performing models, thus associated with a lower value of $\mu$, also have a more variable performance and thus a larger value of $\sigma$). Results are displayed in Table~\ref{gaussian}. 

Comparing the gray column in Table~\ref{gaussian} to Table~\ref{tab:Stats_Hippo_110}, one can observe that both the standard error and the confidence interval width are close to the previously obtained experimental values (for $k \leq 100)$). Nevertheless, the experimental values are slightly lower than those of the simulation. As the value of $k$ increases, the gap between experimental and simulated values decreases.

\section{Discussion}
In this paper, we have provided elements regarding the precision with which segmentation performance can be estimated in typical medical imaging studies. As hypothesized, the test set size needed to obtain a given confidence interval is smaller than in image classification. 
Typically, with 10,000 samples for classification (for a high performing model with over 90\% accuracy), one obtains a $1.2\%$-wide CI. For our segmentation experiments, such width is obtain with only about 1000 samples. For a 4\%-wide CI, one needs about 1000 samples for the aforementioned classification and about 100 samples for segmentation. Of course, one needs to keep in mind that this depends not only on the test sample size but also on $\sigma$ of the performance. For example for $\sigma$ = 5, the 1\%-wide CI would then require between 300 and 500 samples.

Confidence intervals are rarely reported in medical image segmentation papers. To illustrate this, we have conducted a search for  papers published in 2022 in {\sl IEEE Transactions on Medical Imaging (IEEE TMI)} dealing with  segmentation of 3D images.  We found 51 papers containing 107 experiments (a given paper may include several experiments).  Only  71 (66\%) of these experiments (corresponding to 21 papers) were done on an independent test set, the rest reporting cross-validation results on training/validation sets. This is problematic for two reasons. First, the performance on cross-validation results can lead to optimistically biased performances~\cite{wen2020convolutional}. Moreover, they cannot be used to estimate standard errors. For the experiments that had an independent test set, the median size of the test set was 21 (minimum: 4, mean: 61.5, maximum: 697). In light of our experiments, it is not unreasonable that the associated confidence intervals will often be wide (typically between 9 and 10 for a test set with 20 samples). Finally, of the 21 papers that adopted an independent test set, only 3 of them (amounting to 6\% of the 51 surveyed papers) reported confidence intervals or standard-error~\cite{zhao_segmentation_2022, chen_beyond_2022,han_deep_2022}. Note that 11 other papers used statistical testing to compare different approaches even though they did not provide confidence intervals~\cite{song_global_2022,chen_attention-assisted_2022,jiang_unpaired_2022,hung_cat-net_2022, mehta_propagating_2022, huang_3d_2022, chen_semi-supervised_2022,valanarasu_kiu-net_2022, zheng_automatic_2022, wang_recistsup_2022, chen_model-driven_2022}.  

Our study has the following limitations. First, we studied only one dataset. Second, we used only one segmentation model. Future work would need to assess other models and datasets, in particular to see which is the typical range of values that can be expected for $\sigma$. Second, it was restricted to the Dice performance metric and it would be interesting to study if the same observations hold for other metrics (e.g. Hausdorff distance, volume error\ldots). 

Overall, the experiments presented in our paper show the importance of reporting confidence intervals on independent test sets and that, in general, studies with small test sets cannot claim accurate estimation of the performances. We believe that it is an important issue on which the community should focus.

\section{Compliance with ethical standards}
This research study was conducted retrospectively using human subject data made available in open access by the Medical Segmentation Decathlon challenge~\cite{decathlon_short,simpson2019large}. Ethical approval was not required as confirmed by the license attached with the open access data.

\section{Acknowledgments}
\label{sec:acknowledgments}
{\
The research leading to these results has received funding from the French government under management of Agence Nationale de la Recherche as part of the "Investissements d'avenir" program, reference ANR-19-P3IA-0001 (PRAIRIE 3IA Institute) and reference ANR-10-IAIHU-06 (Agence Nationale de la Recherche-10-IA Institut Hospitalo-Universitaire-6).

\bibliographystyle{IEEEbib}
\footnotesize{
    \bibliography{strings,refs, references}
}

\end{document}